\documentclass{article}
\usepackage{spconf,amsmath,graphicx,xtab,longtable,multirow}
\usepackage{booktabs}
\usepackage{multirow}
\usepackage{amssymb}
\usepackage[normalem]{ulem}
\useunder{\uline}{\ul}{}
\usepackage{bbm}
\usepackage{threeparttable}
\usepackage{bbding}
\usepackage{subfigure}
\newcommand{\setParDis}{\setlength {\parskip} {-0.1cm} }
\newcommand{\setParDef}{\setlength {\parskip} {0pt} }
\makeatletter
\renewcommand\normalsize{%
	\@setfontsize\normalsize\@xpt\@xiipt
	\abovedisplayskip 1\p@ \@plus2\p@ \@minus5\p@
	\abovedisplayshortskip \z@ \@plus3\p@
	\belowdisplayshortskip 6\p@ \@plus3\p@ \@minus3\p@
	\belowdisplayskip \abovedisplayskip
	\let\@listi\@listI}
\makeatother

\title{Dual-stream contrastive predictive network with joint handcrafted feature view for SAR ship classification}

\name{Xianting Feng$^{*}$, Hao Zheng$^{*}$, Zhigang Hu$^{\S}$, Liu Yang, Meiguang Zheng\thanks{$^{*}$Co-first authors. $^{\S}$Zhigang Hu is the corresponding author.}}
\address{School of Computer Science and Enginneering, Central South University, China}

\begin{document}
%
\maketitle
\begin{abstract}
Most existing synthetic aperture radar (SAR) ship classification technologies heavily rely on correctly labeled data, ignoring the discriminative features of unlabeled SAR ship images. Even though researchers try to enrich CNN-based features by introducing traditional handcrafted features, existing methods easily cause information redundancy and fail to capture the interaction between them. To address these issues, we propose a novel dual-stream contrastive predictive network (DCPNet), which consists of two asymmetric task designs and the false negative sample elimination module. The first task is to construct positive sample pairs, guiding the core encoder to learn more general representations. The second task is to encourage adaptive capture of the correspondence between deep features and handcrated features, achieving knowledge transfer within the model, and effectively improving the redundancy caused by the feature fusion. To increase the separability between clusters, we also design a cluster-level tasks. The experimental results on OpenSARShip and FUSAR-Ship datasets demonstrate the improvement in classification accuracy of supervised models and confirm the capability of learning effective representations of DCPNet.
\end{abstract}

\begin{figure}[htb]

	\begin{minipage}[b]{1.0\linewidth}
		\centering
		\centerline{\includegraphics[width=8.7cm]{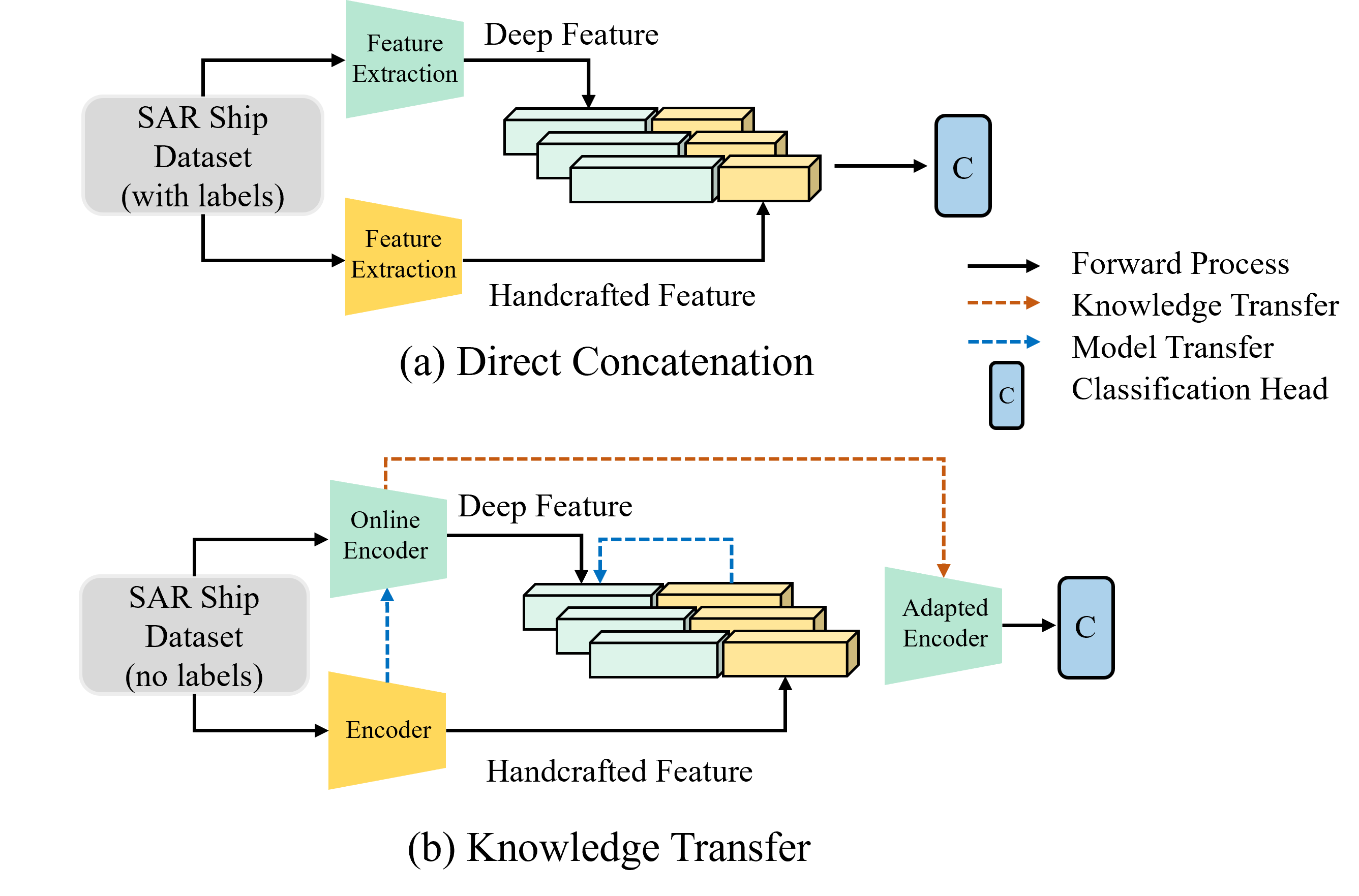}}
	\end{minipage}
%
	
	\caption{Conceptual illustration of different modes about leveraging the handcrafted feature. Method (a) requires labeled data throughout, whereas method (b) is only required for downstream task fine-tuning.}
	\label{pre}


\end{figure}

\begin{figure*}[htp]
	
	\begin{minipage}[b]{1.0\linewidth}
		\centering
		\centerline{\includegraphics[width=15cm]{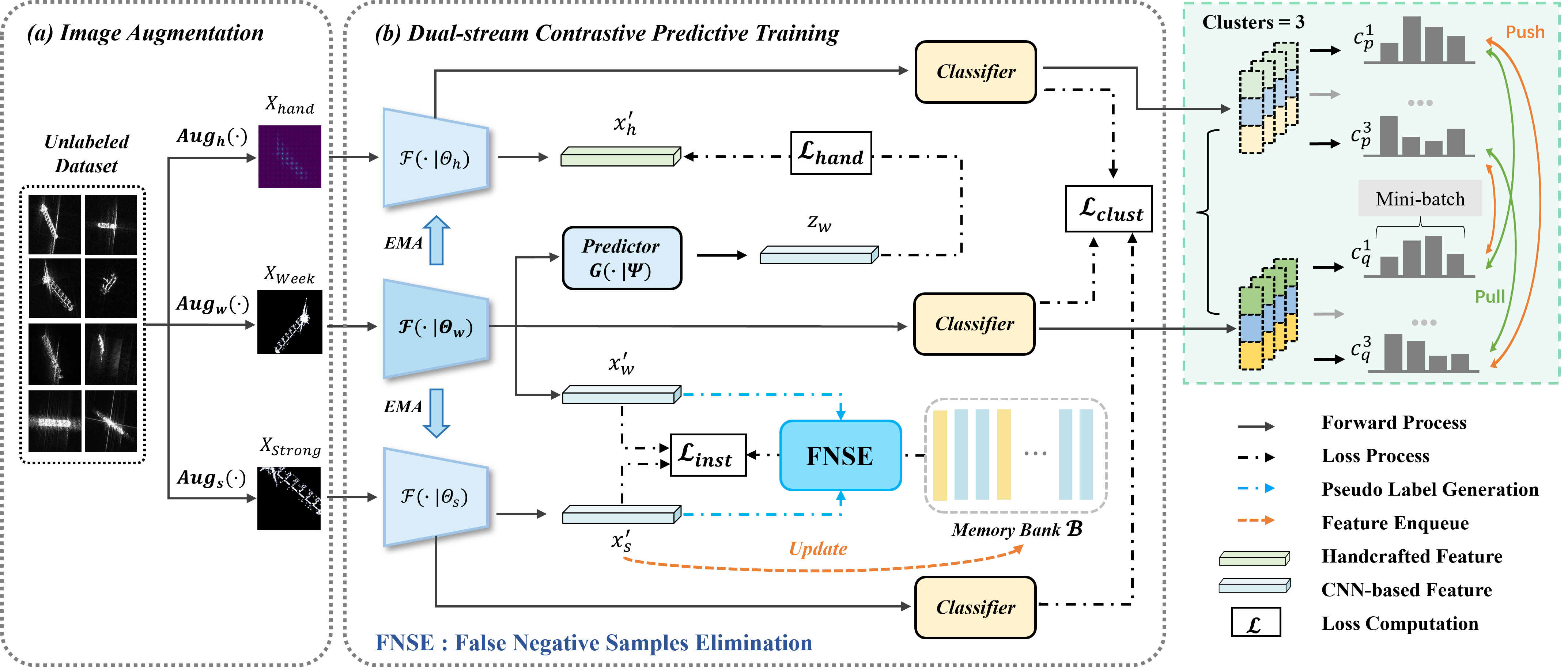}}
	\end{minipage}
	
	\caption{Method overview. Memory bank is updated once after each epoch by collecting the features generating from the encoder $\mathcal{F}\left(\cdot \mid \Theta_{s}\right)$. Pseudo-labels are updated after each epoch according to the confidence scores of both weak and strong view of samples $x_{w}^{'}$, $x_{s}^{'}$ before using it to select the unreliable negative samples. The three losses $\mathcal{L}_{\mathit {hand }}$, $\mathcal{L}_{\mathit {inst }}$ and $\mathcal{L}_{\mathit {clust }}$ are calculated in three separate tasks.}
	\label{frame}
	
\end{figure*}

\section{Introduction}
\label{sec:intro}

Synthetic Aperture Radar (SAR) is an active microwave remote sensing imaging system, whose all-day and all-weather working capacity makes it become the most effective part in ocean applications. As an important ocean mission, SAR ship classification has always been a hot research topic and greatly benefited from the development of deep neural networks during past years \cite{SAR2, SAR3}.


Effective feature extraction plays an important role in image classification task. However, due to the acquisition and annotation costs, it is difficult to collect a large-scale labeled data of remote sensing images, which inevitably affects the richness of embeddings and thus restricting the improvement of model performance. In recent years, traditional handcrafted features have been introduced to alleviate this dilemma. Huang et al. \cite{huang2018multiple} directly combined the Gabor-based MS-CLBP, patch-based MS-CLBP and BOVW-based feature by the multi-feature learning framework. Zhang et al. \cite{ZHANG2022108365} proposed a polarization fusion network with geometric feature embedding to enhance the ship representation. Zheng et al.\cite{MFC} input raw and handcrafted images into two identical backbone networks and performed channel fusion during supervised training. As illustrated in Fig. \ref{pre} (a), the above methods mostly focus on the direct concatenation of high-dimensional features, which not only tends to create redundancy, but also fails to capture the interaction between features.


Inspired by the success of contrastive learning (CL) methods MoCo \cite{MOCO}, BYOL \cite{BYOL}, SimSiam \cite{SIMSIAM} etc., which are capable of learning discriminative features between multiple representations, we utilize CL for the first time to learn complementary information between handcrafted and deep features. As shown in Fig. \ref{pre} (b), handcrafted knowledge can be transferred with the update of model parameters.

Unlike supervised models, the self-supervised pre-training model DCPNet focuses on the connections between samples rather than between samples and labels, which not only enables the full utilization of unlabeled data, but also realizes the reuse of handcrafted knowledge by migrating model to downstream classification tasks.

Specifically, the main contributions of this paper are as follows:
\begin{itemize}
\setParDis
\item A novel dual-stream contrastive predictive network (DCPNet) with a false negative samples elimination (FNSE) module is proposed as a pre-training network and thus obtaining an encoder with good generalization performance for SAR Ship image classification. 
\item In DCPNet, a handcrafted feature branch is designed to guide the transfer of complementary information generating during the process of prediction task.
\item The cluster consistency loss is introduced on the basis of contrastive loss at the instance level, which ensures the separability between samples and the compactness within different categories.
\setParDef
\end{itemize}

\section{METHODOLOGY}
\label{sec:format}

\subsection{Preliminaries}
Considering the labeled SAR ship dataset $\mathcal{X}=\left\{(x_i,\ y_i)\right\}_{i=1}^B$, where $x_{i}$ is the image and $y_{i}$ is the class label, the unlabeled dataset $\mathcal{U}=\left\{\mathit{u}_i\right\}_{i=1}^B$ is fed to network in the pre-training stage. Ultimately our goal is to obtain the online encoder that is updated end-to-end. An overview of DCPNet is shown in Fig. \ref{frame}, which consists of the early image augmentation stage and the core dual-stream contrastive predictive training stage. The specific design principles and training strategies of each part are described in detail below.

\subsection{Image Augmentation Stage}
\label{ssec:subhead2.1}

In this stage, three augmentation processing are designed to obtain different views of training samples, including the handcrafted feature extraction ${Aug}_{h}(\cdot)$ and two deep learning image transformations with different strengths, i.e. week augmentation ${Aug}_{w}(\cdot)$ and strong augmentation ${Aug}_{s}(\cdot)$.

The process of ${Aug}_{h}(\cdot)$ is to introduce physically interpretive information. Many experiments have demonstrated that, the injection of handcrafted feature can complement the deep feature due to its characteristics about being able to focus on the specific physical information \cite{HOG-ShipCLSNet, YUN20121678, CHEVTCHENKO2018748, hou2020fusar}. 

Besides, two transformations of ${Aug}_{w}(\cdot)$ and ${Aug}_{s}(\cdot)$ are raised to simulate the different working conditions. Considering the properties of SAR images, we remove invalid operations from the settings of existing research \cite{Whatshouldnot}. Specifically, ${Aug}_{w}(\cdot)$ concluded simple random cropping, horizontal flips etc., is regarded as an anchor for providing the benchmark to guide the generation of pseudo-labels, while ${Aug}_{s}(\cdot)$ concluded gaussian blur, color jitter etc., is used to enrich the feature bank and thus enhancing the generalization of our model.


\subsection{Dual-stream Contrastive Predictive Training Stage}
\label{ssec:subhead2.2}


CL essentially guides model training by establishing pretext tasks. For example, SimCLR constructs positive and negative pairs through augmentations within mini-batch. MoCo builds a dictionary look-up task by designing a dynamic dictionary with a queue and a moving-averaged encoder. Different from these methods, DCPNet adopts three new pretext tasks for classification from the prespective of feature fusion, including the handcrafted prediction task, the image similarity comparison task and the image cluster consistency task.
\\\textbf{Handcrafted feature prediction tasks.} Considering that handcrafted features are far away from deep features in the embedding space, forcing a comparison of two features at the instance level will affect the focus of feature extraction and make it difficult for the model to converge.

In order to address this problem, the prediction head $G(\cdot \mid \Psi)$ is connected behind the encoder $\mathcal{F}\left(\cdot \mid \Theta_{w}\right)$ to realize the knowledge transfer of handcrafted features within the model by calculating the similarity loss between the output $z_{w}$ of the prediction head and the output $x_{h}$ of the encoder. This loss based on mean square error is defined as follows:

\begin{equation}
	\mathcal{L}_{\mathit {hand }}=2-2 \cdot \frac{\left\langle G\left(x_{w} \mid \Psi\right), x_{h}\right\rangle}{\left\|G\left(x_{w} \mid \Psi\right)\right\|_{2} \cdot\left\|x_{h}\right\|_{2}}
\end{equation}
\\\textbf{Instance-level image similarity comparison task.} For each sample, different augmentations of the same image should be brought "nearby" in an embedding space since they likely contain similar semantic content or correlated features. 


Following the InfoNCE loss \cite{oord2018representation}, strong augmentations are utilized for image to generate a memory bank for storing features. Taking $x_{w}^{i}$ as an anchor, $\left(x_{w}^{i}, x_{s}^{i}\right)$ and $\left(x_{\mathrm{w}}^{i},\left\{k^{j}\right\}_{j=1}^{K^{\prime}}\right)$ are defined respectively as positive pair and a multi-negative pair. In particular, ${\displaystyle{\left\{k^{j}\right\}_{j=1}^{K^{\prime}}}}$ represents the memory bank containing $K^{\prime}$ embeddings after eliminating false negative samples through pseudo-labels, which will be described later. Finally, the instance-level contrastive loss is defined as follows:

\begin{equation}
	\mathcal{L}_{\mathit {inst}}=-\frac{1}{N} \sum_{i=1}^{N} \log \frac{e^{s\left(x_{w}^{i}, x_{s}^{i}\right) / \tau}}{e^{s\left(x_{w}^{i}, x_{s}^{i}\right) / \tau}+\sum_{j=1}^{K^{\prime}} e^{s\left(x_{w}^{i}, k^{j}\right) / \tau}}
\end{equation}

where $s\left(u,v\right)$ is the similarity function, i.e., the inner product $s\left(u, v\right)=u^{T}v$, and $\tau$ is the temperature factor.
\\\textbf{Cluster-level image cluster consistency task.} Instance-level constraint is equally important as cluster-level constraint. The same batch of images should have similar category distributions under different augmentations.
Our framework projects multi-view features into an M-dimensional (M is the number of ship categories) space through a classifier, and utilizes a consistency loss to promote the compactness within classes in feature spaces. Specifically, using $\left(c_{p}^{i}, c_{q}^{i}\right) \in \mathbbm{R}^{N \times 1}$ to denote the distribution representation of cluster $i$ under two augmentation strategies $p$ and $q$, corresponding to the above, $p, q \in\left\{A_{\mathit {weak }}, A_{\mathit {strong }}, A_{\mathit {handcrafted }}\right\}$. And the loss is defined as follows:  

\begin{equation}
	\mathcal{L}_{\mathit {clust}}=-\frac{1}{M} \sum_{i=1}^{M} \log \frac{e^{s\left(c_{p}^{i}, c_{q}^{i}\right) / \tau}}{e^{s\left(c_{p}^{i}, c_{q}^{i}\right) / \tau}+\sum_{j=1}^{M} \mathbbm{1}_{[i \neq j]} e^{s\left(c_{p}^{i}, c_{q}^{j}\right) / \tau}}
\end{equation}

where $s\left(u,v\right)$ represents the $L_{2}$ normalized cosine similarity $sim\left(u, v\right)=u^{T} v /\|u\|\|v\|$, and $\mathbbm{1}_{\left[i\neq j\right]}$ represents distribution representation that does not belong to the same ship category. Finally, three training objectives are minimized to train the core encoder $\mathcal{F}\left(\cdot \mid \boldsymbol{\Theta}_{w}\right)$. All of them simultaneously improve the quality of feature representations and classifiers. The overall loss should be:

\begin{equation}
	\mathcal{L_{\mathit {overall }}}=\alpha \mathcal{L}_{\mathit {hand }}+\beta \mathcal{L}_{\mathit {inst }}+\gamma \mathcal{L}_{\text {clust }}
\end{equation}

where the loss coefficients $\alpha,\beta,\gamma$ satisfy $\alpha+\beta+\gamma=1$.
\\\textbf{False negative sample elimination.} Generally, negative pairs are formed by sampling views from different images without labels, which may ignore the semantic information within them, and thus resulting in unreliable negative samples in the memory bank \cite{BoostingCL}. Therefore, pseudo-labels are utilized to weaken the impact of false negative samples on model training. Define $c$ as the degree of confidence, the pseudo-label probability vector can be expressed as:

\begin{equation}
	\widehat{\mathcal{P}}_{i}^{\mathit {elim }}=c \cdot \widehat{\mathcal{P}}_{i}^{w}+(1-c) \cdot \widehat{\mathcal{P}}_{i}^{s}
\end{equation}

According to FixMatch \cite{Fixmatch}, we use confidence to filter out labels with high reliability for elimination and discard the low-confidence pseudo-labels. 
The new queue after eliminating false negative samples based on sample $i$ is as follows:

\begin{equation}
	\left\{k^{j}\right\}_{j=1}^{K^{\prime}}=\left\{k_{orig}^{j}\right\}_{j=1}^{K} \cdot \mathit{mask}^{p} \cap \mathbbm{1}_{\hat{\mathcal{P}}_{i} \neq \hat{\mathcal{P}}_{j}}
\end{equation}

where $\left\{k_{orig}^{j}\right\}_{j=1}^{K}$ is the original negative sample queue, and $\mathit{mask}^{p}$ represents the filter strategy for retaining pseudo-labels with a confidence level above the setting threshold.

\begin{table*}[htp]
	\centering
	\renewcommand\arraystretch{1.1}  
	\caption{Evaluation accuracy($\%$) of applying two evaluation method "Fine-tuning" and "KNN-way" for DCPNet and other advanced self-supervised methods.}
	\label{tableda}
	\vspace{1.4mm} 
	\scalebox{1.0}{
		\begin{threeparttable}
		\begin{tabular}{@{}cccccccc@{}}
			\hline
			\hline
			\multirow{2}{*}{Dataset}     & \multirow{2}{*}{Method} & \multirow{2}{*}{Pre-training ep.} & \multicolumn{3}{c}{Fine-tuning}                                                   & \multicolumn{2}{c}{KNN-way}                           \\ \cline{4-8}
			&                                  &                                & Ft1.                      & \multicolumn{1}{c}{Ft2.}  & Ftall.                    & ResNet18                        & ResNet50                        \\
			\hline
			\multirow{5}{*}{OpenSARShip} & MoCo                             & 200                            & 66.05±1.74                & \underline{71.50±0.47}                & 70.33±1.21                & 67.67±0.95                & 67.85±0.35                \\
			& BYOL                             & 200                            & \underline{68.44±0.29}                & 68.84±1.80                & \textbf{73.03±0.69}                & -                & -                \\
			& SimSiam                          & 200                            & 66.64±1.04                & 66.91±1.55                & \underline{72.79±0.90}                & 68.78±3.02                & 66.64±1.04                \\
			& DCPNet(Ours)                     & 20                             & 60.61±2.57                & 65.68±2.65                & 67.92±3.05                & \underline{69.29±0.29}                & \underline{69.92±0.95}                \\
			& DCPNet(Ours)                     & 200                            & \textbf{70.37±1.01} 		 & \textbf{73.66±1.01}       & 71.50±1.39       & \textbf{70.66±0.42}       & \textbf{69.94±0.60} \\ 
			\hline
			\hline
			\multirow{5}{*}{FUSAR-Ship}   & MoCo                             & 200                            & 57.73±1.22                & \underline{74.94±0.60}                & 74.69±0.91                & \underline{70.13±0.58}                & \textbf{70.29±1.01}                \\
			& BYOL                             & 200                            & \underline{60.21±0.74}                & 73.62±1.40                & \underline{85.25±0.39}                & -                & -                \\
			& SimSiam                          & 200                            & 56.29±1.37                & 61.62±0.89                & 80.71±0.31                & 65.92±0.85                & 56.67±0.35                \\
			& DCPNet(Ours)                     & 20                             & 56.59±0.83                & 72.66±0.48                & 72.49±0.88                & 61.08±0.99                & 55.97±0.66               \\
			& DCPNet(Ours)                     & 200                            & \textbf{69.33±1.62}		 & \textbf{83.18±0.35} 		 & \textbf{87.94±0.76} 		 & \textbf{72.33±2.10}       & \underline{66.67±0.54} \\ 
			\hline
			\hline
		\end{tabular}
		\begin{tablenotes} 
			\footnotesize     
			\item[*] The classification head in Ft1. and Ftall. is a layer of MLP, and in Ft2. is a projector head with two linear layers, a BN layer and a ReLU layer. Ft1. and Ft2. only update classification head during training process, while Ftall. updates all parameters. Besides, the bold black numbers indicate the highest accuracy achieved under different evaluation methods
		\end{tablenotes} 
	\end{threeparttable}
	}
\end{table*}

\section{EXPERIMENT}
\label{sec:typestyle}

\subsection{Experiment Preparation}
\label{ssec:subhead3.1}

\textbf{Dataset.} 1) OpenSARShip: The OpenSARShip dataset is collected by the Senti-nel-1A satellite, which contains three types: bulk carriers , container ships and tankers. 2) FUSAR-Ship: The FUSAR-Ship dataset is extracted from the FUSAR GF-3 SAR dataset, including seven types: bulk carriers, container, fishing, tankers, general cargo, other cargo and others. It has higher resolution compared with the OpenSARShip.
\\\textbf{Implementation Details.} We randomly selected 2,296 and 11,884 ship chips from OpenSARShip and FUSAR-Ship as the pre-training dataset and cropped them to 224×224 size. The experiment consists of two stages. In pre-training stage, the whole network including the encoder(s), projector(s) and predictor(s) is trained in self-supervised way, and the hyperparameters related to CL are following the \cite{MOCO}. The value of three loss coefficients are: $\alpha=0.2$, $\beta=0.6$, $\gamma=0.2$. In evaluation stage, two methods are applied to measure the effectiveness of the detached encoder. One is fine-tuning where a classification head is attached to the encoder and all of them are trained in a supervised manner. The other is KNN classification with memory bank, where k is set to 45. Besides the backbone uses ResNet 18/34/50 in both stages.

%

\subsection{Experiment Results}
\label{ssec:subhead3.2}
\subsubsection{Comparison with other self-supervised methods}
\label{sssec:subsubhead3.2.1}

Table \ref{tableda} shows the evaluation accuracy of DCPNet and the existing CL framework on the SAR ship classification task. It can be seen that when pre-training epochs is set to 200 and using the fine-tuning method, DCPNet has a larger gain than all the CL frameworks except using the "ftall". The reason is that when the test accuracy of “KNN-way” is close to supervised models, extracted features are distinguishable enough to be classified through “ft1./ft2.”. So compared with BYOL and SimSiam, fine-tuning with several layers is more effective.
Secondly, the results of 20 epochs emphasize that DCPNet can approach the CL models with small epochs on openSARShip with “KNN-way” and FUSAR-Ship with “fine-tuning.
Thirdly, when using knn method, although the effectiveness is weaker than fine-tuning, it can also achieve the best results on both datasets while using ResNet18 as backbone.

\subsubsection{Comparison with supervised learning baseline}
\label{sssec:subsubhead3.2.2}
\setlength{\abovecaptionskip}{0.cm}
\setlength{\abovecaptionskip}{-0.cm}

\begin{table}[t]
	\centering 
	\renewcommand\arraystretch{1.08}  
	\caption{Fine-tuning accuracy$($\%$)$ of DCPNet and state-of-the-art supervised learning baseline.}
	\vspace{0.4mm} 
	\label{tablex}
	\scalebox{0.95}{
		\begin{tabular}{c c c c}
			\hline
			\hline
			Method	    				& Train ep.			& OpenSARShip					& FUSAR-Ship\\		
			\midrule
			ResNet-18	    			& 100	           	& 71.70±0.94	    			& 79.71±0.73\\
			ResNet-34	    			& 100			   	& 71.84±1.23	    			& 80.73±0.41\\
			ResNet-50	    			& 100	           	& 72.15±1.20	   				& 80.96±0.47\\
			DCPNet(Best)		& 100	           	& \textbf{73.66±1.01}	    	& \textbf{87.94±0.76}\\
			\hline
			\hline
		\end{tabular}
	}
\end{table}

To further prove the superiority of our framework, the DCPNet framework is compared with baseline supervised models. As shown in Table \ref{tablex}, through our pre-training framework and supervised training with labeled data, the accuracy of benchmark neural networks such as ResNet-18 on downstream tasks increased by 1.51$\%$ and 6.98$\%$ respectively.
Two evaluation methods respectively prove that DCPNet can not only train an encoder that learns prior physical knowledge of handcrafted features without causing redundancy, but also obtain a feature set with better generalization and discrimination. Both the encoder and feature set can demonstrate the excellent performance of DCPNet in downstream tasks.

\section{Conclusion}
\label{sec:conclusion}

This paper proposes a new dual-stream contrastive predictive network (DCPNet) based on the fusion of deep and handcrafted feature for SAR ship classification. The two main contrastive pretext tasks, along with the cluster-level task are designed to learn the inherent general features of images under different augmentations and to achieve collaboration of model parameter update and handcrafted knowledge transfer. Furthermore, this framework enhances the separability between classes by adding the cluster loss.
In addition, pseudo-labels based on confidence are used to filter the memory bank, which improves the effectiveness of the negative samples and corrects the embedding space.
Through two-stage comparative experiments, it is concluded that the performance of the proposed pre-training framework DCPNet in SAR ship classification tasks is significantly higher than the existing CL methods, and the classification accuracy of the supervised benchmark models are also effectively improved. The proposed DCPNet only achieves knowledge transfer for single handcrafted feature, but further research is needed on how to aggregate information between multiple handcrafted features. 
\vfill\pagebreak

%


\bibliographystyle{IEEEbib}
\bibliography{refer}

\clearpage

\appendix

\section{Ablation experiments} 

Three ablation studies are conducted to investigate the effectiveness of each component of the proposed DCPNet. To be specific, the effectiveness of the handcrfted prediction task, the image cluster consistency task and the false negative sample elimination module. In addition, we provide specific analysis of the experimental results as follows.

\subsection{Ablation Study on the Handcrafted Prediction Task}
\label{ssec:subheadA.1}

In the ablation study on traditional handcrafted feature prediction task, we choose HOG feature for research and two sets of experiments are arranged, Experiment 1 is used to explore the contrastive mechanism between handcrafted and deep features to match the distribution of both in the embedding space and Experiment 2 is used to further validate the effectiveness of handcrafted feature knowledge.

\begin{table}[h]
	\centering 
	\renewcommand\arraystretch{1.2}  
	\caption{Contrastive mechanism between handcrafted and deep features}
	\vspace{1.4mm} 
	\label{albtable3}	
	\tabcolsep=0.15cm
	\scalebox{0.95}{
		\begin{tabular}{ccc}
			\hline
			\hline
			Datasets & Contrastive mechanism & Accuracy(\%)\\
			\hline
			\hline
			
			\multirow{2}{*}{FUSAR-Ship}
			& Directly Contrast & 64.90±2.31\\
			& \textbf{Features Prediction(Ours)} & \textbf{87.94±0.76}\\
			
			\hline
			\hline
		\end{tabular}
	}
\end{table}

As shown in Table \ref{albtable3}, with the addition of handcrafted features, the accuracy of direct contrast mechanism drops severely but feature prediction mechanism is greatly improved. 

\begin{figure}[h]
	\centering
	\vspace{-0.1cm}
	\subfigure[Contrast on Weak and Strong Augmentation]{
		\label{aaaa}
		\includegraphics[width=2.7cm]{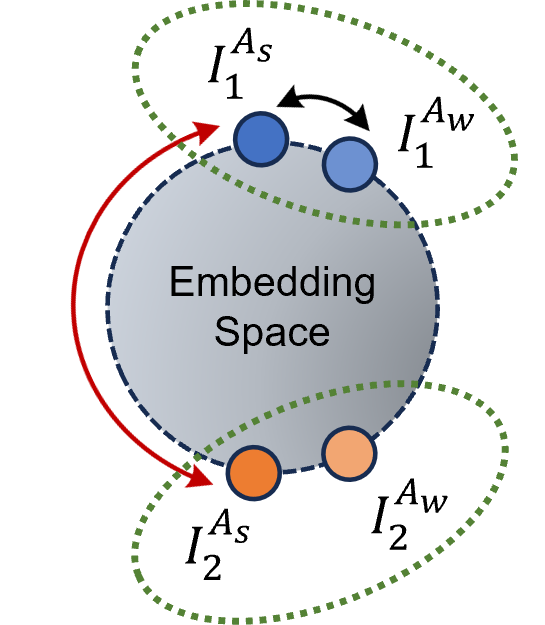}	}
	\quad
	\subfigure[Contrast on Hand-crafted and Weak Augmentation]{
		\label{bbbb}
		\includegraphics[width=3.4cm]{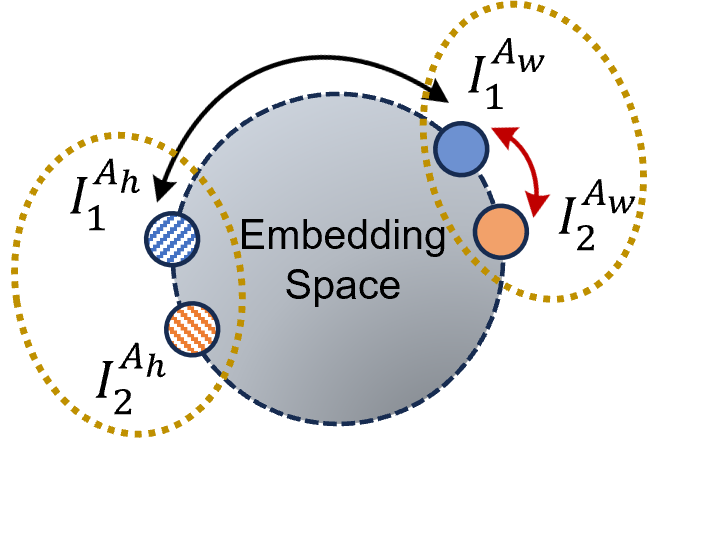}	}\vskip -8pt
	\quad
	\caption{Conceptual illustration of space distribution guided by three augmentation strategies.}
	\label{abl2}
\end{figure}

In Fig. \ref{abl2}, two colors represent two different images respectively, and red arrows indicate pushing away while black arrows indicate pulling closer. It can be seen that two augmentations of the same image are closer, conversely the handcrafted features of images are farther away from their augmentation pattern, which also explains our experimental results in Table \ref{albtable3}.

\begin{table}[h]
	\centering 
	\renewcommand\arraystretch{1.08}  
	\caption{Ablation study on the Handcrafted Prediction Task}
	\vspace{1.4mm} 
	\label{albtable4}	
	\tabcolsep=0.4cm
	\scalebox{0.95}{
		\begin{tabular}{ccc}
			\hline\\[-4.0mm]\hline
			Datasets & Handcrafted Task & Accuracy(\%)\\
		    \hline
		    \hline
			
			\multirow{2}{*}{OpenSARShip}
			
			& \XSolidBrush & 71.70±0.90\\
			& \Checkmark & \textbf{73.66±1.01}\\
			\hline
			\multirow{2}{*}{FUSAR-Ship}
			& \XSolidBrush & 81.10±0.56\\
			& \Checkmark & \textbf{87.94±0.76}\\
			
			\hline
			\hline
		\end{tabular}
	}
\end{table}

Table \ref{albtable4} shows the accuracy results with and without the prediction task. From Table \ref{albtable4}, it can be seen that the utilization of HOG knowledge proves its effectiveness with a huge improvement in accuracy, i.e., 1.96\% on OpenSARShip and 6.84\% on FUSAR-Ship.

\subsection{Ablation Study on the Image Cluster Consistency Task}
\label{ssec:subheadA.2}

Table \ref{table5} shows the accuracy results using constraints of cluster level and not using it. From Table \ref{table5}, the accuracy is improved by 1.10\% and 3.37\%. on OpenSARShip and FUSARShip respectively, because this task guides the model to learn more fine-grained distinguishable information of cluster.

\begin{table}[h]
	\centering 
	\renewcommand\arraystretch{1.10}  
	\caption{Ablation study on the Cluster Consistency Task}
	\vspace{1.4mm} 
	\label{table5}
	\tabcolsep=0.4cm
	\scalebox{0.95}{
		\begin{tabular}{ccc}
			\hline\\[-4.0mm]\hline
			Datasets & cluster Task & Accuracy(\%)\\
			\hline
			\hline
			
			\multirow{2}{*}{OpenSARShip}
			
			& \XSolidBrush & 72.56±0.22\\
			& \Checkmark & \textbf{73.66±1.01}\\
			\hline
			\multirow{2}{*}{FUSAR-Ship}
			& \XSolidBrush & 84.57±0.59\\
			& \Checkmark & \textbf{87.94±0.76}\\
			
			\hline
			\hline
		\end{tabular}
	}
\end{table}

\subsection{Ablation Study on the False Negative Sample Elimination Module}
\label{ssec:subheadA.3}

Table \ref{table6} shows the accuracy results retaining the false negative samples and eliminating them.
From Table \ref{table6}, the accuracy gain of 3.00\% was achieved on OpenSARShip and 4.55\% on FUSAR-Ship. These results demonstrate that correct negative samples lead the model to converge better since the proposed method is based on the comparison of sample pairs.
\begin{table}[h]
	\centering 
	\renewcommand\arraystretch{1.15}  
	\caption{Ablation study on the False Negative Sample Elimination(FSNE) Module}
	\vspace{1.4mm} 
	\label{table6}
	
	\tabcolsep=0.65cm
	\scalebox{0.95}{
		\begin{tabular}{ccc}
			\hline\\[-4.0mm]\hline
			Datasets & FSNE & Accuracy(\%)\\
			\hline
			\hline
			
			\multirow{2}{*}{OpenSARShip}
			
			& \XSolidBrush & 70.66±0.42\\
			& \Checkmark & \textbf{73.66±1.01}\\
			\hline
			\multirow{2}{*}{FUSAR-Ship}
			& \XSolidBrush & 83.39±0.26\\
			& \Checkmark & \textbf{87.94±0.76}\\
			
			\hline
			\hline
		\end{tabular}
	}
\end{table}

\end{document}